\documentclass[10pt,twocolumn,letterpaper]{article}

\usepackage{wacv}
\makeatletter
\@namedef{ver@everyshi.sty}{}

\usepackage{times}
\usepackage{epsfig}
\usepackage{graphicx}
\usepackage{amsmath}
\usepackage{amssymb}
\usepackage{booktabs}

\usepackage[accsupp]{axessibility} 
\usepackage{subcaption}
\usepackage{tikz}
\usepackage{cite}
\usepackage{multirow}
\usepackage{subcaption}
\usepackage{xcolor}
\usepackage{color, colortbl}

\def\wacvPaperID{2120} 

\wacvapplicationstrack 

\wacvfinalcopy 

\definecolor{thesis-lightcyan}{RGB}{200,255,255}
\newcommand{\mycc}{\cellcolor{thesis-lightcyan}}

\newcommand{\ourmethod}{\texttt{LAB}}
\newcommand{\ournetwork}{\texttt{LAB-BNN}}
\newcommand{\sign}{\texttt{sign(.)}}
\newcommand{\ourmetric}{ENDSIM}

\ifwacvfinal
\usepackage[breaklinks=true,bookmarks=false]{hyperref}
\else
\usepackage[pagebackref=true,breaklinks=true,colorlinks,bookmarks=false]{hyperref}
\fi

\begin{document}

\title{LAB: Learnable Activation Binarizer for Binary Neural Networks}

\author{Sieger Falkena$^{1,2}$, Hadi Jamali-Rad$^{1,2}$, Jan van Gemert$^{1}$\\
$^{1}$ TU Delft, Delft, The Netherlands \\ 
$^{2}$ Shell Global Solutions International B.V., Amsterdam, The Netherlands\\
{\tt\small sieger.falkena@shell.com, h.jamalirad@tudelft.nl, j.c.vangemert@tudelft.nl}}

\setcounter{page}{123}

\maketitle
\thispagestyle{empty}

\begin{abstract}
\vspace{-4mm}
Binary Neural Networks (BNNs) are receiving an upsurge of attention for bringing power-hungry deep learning towards edge devices. The traditional wisdom in this space is to employ \sign{} for binarizing feature maps. We argue and illustrate that \sign{} is a \emph{uniqueness bottleneck}, limiting information propagation throughout the network. To alleviate this, we propose to dispense \sign{}, replacing it with a learnable activation binarizer (\ourmethod), allowing the network to learn a fine-grained binarization kernel per layer - as opposed to global thresholding. \ourmethod{} is a novel universal module that can seamlessly be integrated into existing architectures. To confirm this, we plug it into four seminal BNNs and show a considerable accuracy boost at the cost of tolerable increase in delay and complexity. Finally, we build an end-to-end BNN (coined as \ournetwork) around \ourmethod, and demonstrate that it achieves competitive performance on par with the state-of-the-art on ImageNet. Our code can be found in our repository: \url{https://github.com/sfalkena/LAB} \footnote{This paper is accepted to appear in the proceedings of WACV 2023}.

\end{abstract}
\vspace{-6mm}
\section{Introduction}
\label{sec:intro}
Convolutional Neural Networks (CNNs) dominate the current state-of-the-art computer vision tasks. With evolving research, models gained increasingly higher accuracy, but in parallel they have grown in size and complexity. This imposes a significant burden for deploying deep learning models on resource-constrained edge devices. Recent studies explore model compression techniques to reduce model size and latency, such as pruning \cite{liu2018rethinking}, quantization \cite{xu2018deep}, knowledge distillation \cite{gou2021knowledge}, neural architecture search \cite{elsken2019neural} and low rank approximation \cite{yu2017compressing}. The most extreme form of quantization is realized by binarization, resulting in binary weights and activations $\{-1, +1\}$. Networks utilizing this are known as Binary Neural Networks (BNNs) and promise a bright future for energy-efficient deep learning. By quantizing weights and activations aggressively, one can theoretically achieve a memory reduction of $32\times$ and a computational speedup of $58\times$ on typical CPUs \cite{rastegari2016xnor}.

\begin{figure*}[t!]
\centering
   \includegraphics[width=\linewidth]{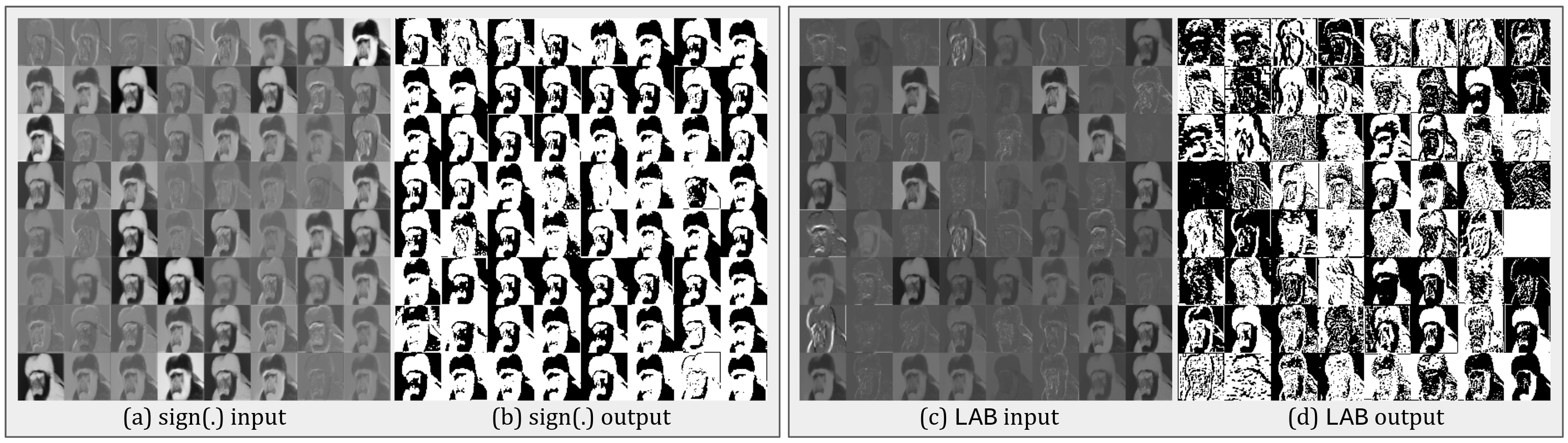}
   \caption{All channels before and after the binarization step for the first binary convolution after model training. In contrast to \sign, binarization using \ourmethod{} does not map some of the discrete output feature maps into similar binary feature maps, but learns to distinguish important features during binarization, improving the information propagation capacity of the BNN.}
\label{fig:teaser}
\end{figure*}

\newpage
The current consensus in literature is to use \sign{} as a mapping from the full-precision to binary values. However, this imposes three widely-known issues: (i) the representational power of \sign{} with respect to the floating point counterpart decreases from $2^{32}$ to only $2$ information levels \cite{liu2018bi}; (ii) the derivative of \sign{} is a Dirac Delta returning a zero gradient almost everywhere \cite{bulat2019improved}; (iii) \sign{} spatially uses the same threshold everywhere, which we further refer to as the global threshold. In this work, we identify and debate about a fourth problem (iv) which we refer to as \emph{uniqueness bottleneck}. Approaching the problem from both qualitative and quantitative angles, we demonstrate that using \sign{} further limits the representational capacity of the network. Interestingly, several studies state that \sign{} is a sub-optimal binarization operation and that it is not straightforward to find a new binarization function \cite{bulat2020high, wang2020sparsity}. This is exactly why we embark on this challenge in this paper.

Multiple remedies have been proposed to cope with the aforementioned issues of \sign{}, including the introduction of scaling factors \cite{rastegari2016xnor}, gradient approximation \cite{qin2020forward} and pre-binarization distribution shaping \cite{kim2021improving}. Amongst these directions, we believe that the pre-binarization reshaping shows the most potential to alleviate the information bottleneck of BNNs. However, in contrast to the existing studies, we argue that shaping the pre-binarization distribution is a means an the end, and not the end in itself. To address the issues enumerated, we design a \emph{learnable} activation binarization function (\ourmethod) to automate the mapping from the full-precision feature maps to the binary counterparts, so that the representational capacity of the network (compared to full-precision) is least impacted. This is shown schematically in Fig.~\ref{fig:teaser} (and elaborated in Section~\ref{sec:problem}), where application of a global \sign{} threshold on the diverse spectrum of values in the discrete feature maps results in similar looking outputs (acting like a diversity bottleneck), whereas \ourmethod{} can potentially avoid such loss of information and reveal important features for later layers.
\newpage
Our contributions can be summarized as follows:
\vspace{-1mm}
\begin{itemize}
    \item To the best of our knowledge, for the first time, we identify the \emph{uniqueness bottleneck} imposed by the \sign{}. We demonstrate that \sign{} limits the representational capacity of binary feature maps.
    \item To address this bottleneck, we introduce a novel learnable activation binarization: \ourmethod. We show that \ourmethod{} is a universal module that can readily be plugged into any existing BNN architecture, and improve its performance. Our experimentation on four seminal BNN baselines corroborates this claim.  
    \item We build an end-to-end network around \ourmethod{} (coined as \ournetwork) and demonstrate that it offers competitive accuracy ($64.2\%$ Top-$1$ validation accuracy) on par with the state-of-the-art in this space on ImageNet.  
\end{itemize}

\section{Related Work}
Current BNNs binarize the full precision weights and activations by applying \sign{} on them: 
\vspace{-1mm}
\begin{equation}
\label{eq:sign}
x_{b}=\operatorname{\texttt{sign}}\left(x_{r}\right)=\left\{\begin{array}{l}
+1, \text { if } x_{r}>0 \\
-1, \text { if } x_{r} \leq 0
\end{array}\right.,
\end{equation}
where $x_b$ and $x_r$ denote the binary and real (full precision) values, respectively. Naively applying these quantizations to a CNN yields low accuracy and to close the gap between BNN implementations and their real-valued counterparts, several research directions have arisen: minimization of quantization error \cite{bulat2019improved, rastegari2016xnor}, loss function improvement \cite{liu2020reactnet, ding2019regularizing, hou2016loss}, gradient approximation \cite{liu2020reactnet, lin2020rotated, qin2020forward}, different network architecture designs \cite{liu2018bi, liu2020reactnet, howard2017mobilenets, bulat2020bats, zhu2020nasb, chen2021bnn}, training strategies \cite{martinez2019training, alizadeh2018empirical, liu2021adam, tang2017train} and binary inference engines \cite{bannink2021larq, fromm2020riptide, yang2017bmxnet, zhang2019dabnn}. Apart from these main directions, a few studies investigate new methodologies for binarizing weights and activations, which will be elucidated next. 

\textbf{Weight binarization.} A novel approach for weight binarization is presented in \cite{han2020training} where both full precision and binary weights are employed as noisy supervisors for learning a mapping towards the final binary weights. As this mapping is learnable, it can exploit the relationships between weights. SiMaN \cite{lin2021siman} and RBNN \cite{lin2020rotated} both propose a new binarization method based on so-called angle alignment between the full-precision and binary weights. 

\textbf{Activation binarization.} One way to approach activation binarization is through classic computer vision techniques, such as dithering. This technique can binarize an image in a way that shifts quantization error towards higher frequencies. As the human visual system is more receptive to lower frequencies, the binarized image is perceived as having a low quantization error, and thus, carrying more information. A realization of this idea is called DitherNN but it only reports mild improvement \cite{ando2018dither}. Most activation binarization approaches focus on shaping the pre-binarization distribution, from which a higher entropy can be achieved after binarization \cite{qin2020forward}. For instance, an extra regularization term is proposed in \cite{ding2019regularizing} to explicitly shape the pre-binarization distribution so that it counteracts degeneration, saturation, and gradient mismatch problems. It is argued in \cite{kim2021improving} that BNNs benefit from an unbalanced pre-binarization distribution. ReActNet \cite{liu2020reactnet} argues that BNNs benefit from learning a similar activation distribution as their full-precision counterparts. On a related note, Si-BNN \cite{wang2020sparsity} approaches the activation binarization problem from a somewhat different angle and introduces sparsity in the activation binarization process. Even though these studies show promising performance results for BNNs, we argue that changing the pre-binarization distribution is still a form of adaptive global thresholding, and thus a sub-optimal approach. Therefore, the output feature map does not fully reflect on, and adapt to the local information of the input feature map. 

\section{Problem Formulation}
\label{sec:problem}
\vspace{-2mm}
In this section, we reflect on the limitation a \sign{} binarization function inflicts on a BNN. We examine the capability of the network to cope with the single threshold value of \sign, which we refer to as \emph{global} thresholding - given that its value is the same for all spatial dimensions. In practice, as the input feature map to \sign{} is the output of a previous (convolutional) layer, the kernel of that convolution will learn to push or pull parts of the output feature map above or beneath the global threshold value, resulting in the fact that the output feature map will be closely centered around the threshold value. The batch normalization layer can further guide this process by effectively shifting the threshold value. Although this combination is essential for the current learning process of BNNs, we argue that there still is a limitation in its efficacy. 

\begin{figure*}[t!]
\centering
\includegraphics[width=0.6\linewidth]{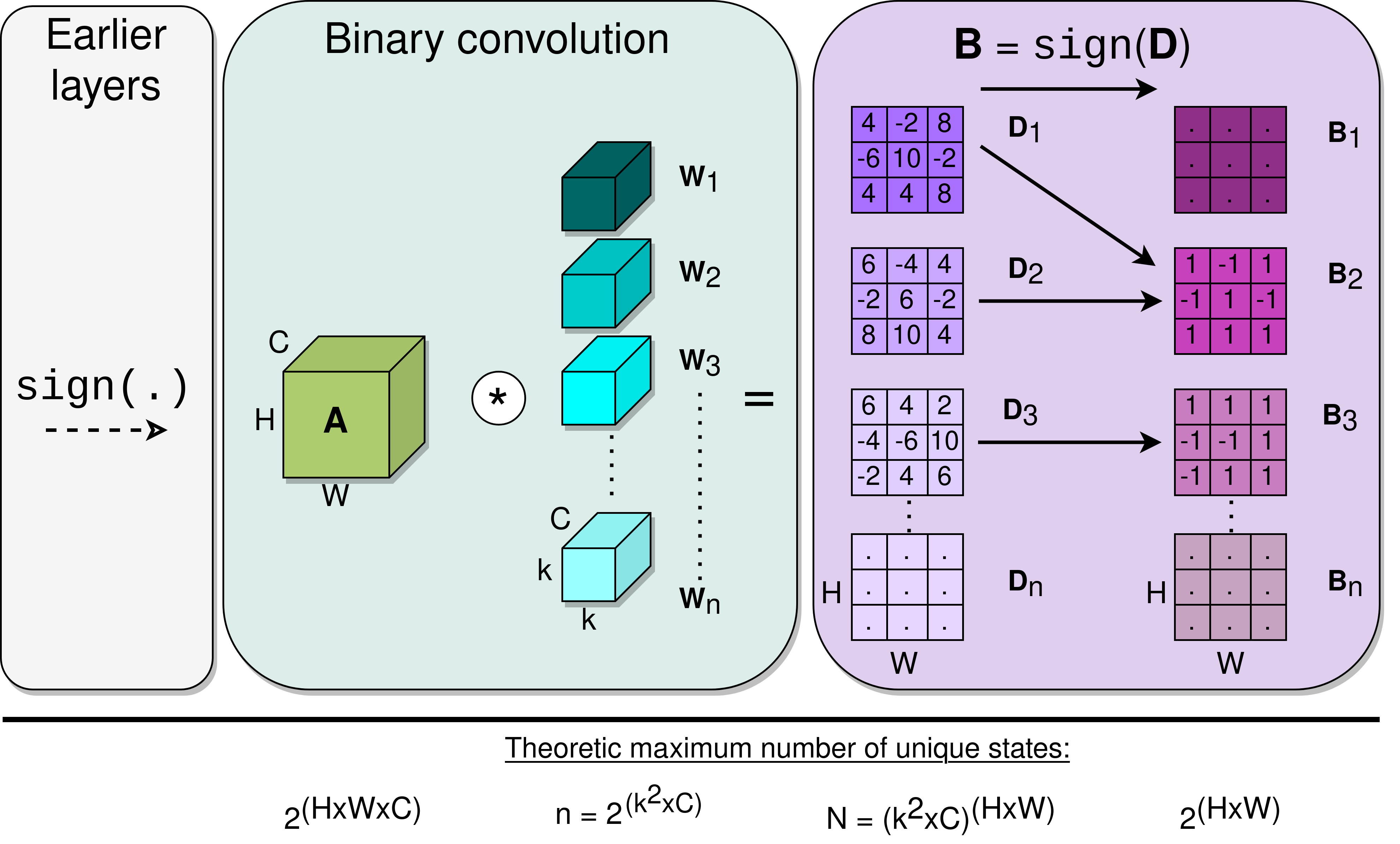}
\caption{The \emph{uniqueness bottleneck}. Activation ${\bf A}$ is convolved with all unique kernels ${\bf W}_i$'s. The finite-alphabet feature maps ${\bf D}$ are binarized by the \sign, which creates the bottleneck of multiple ${\bf D}$'s mapping to the same binary feature map ${\bf B}$. The equations (\textit{at the bottom}) indicate the theoretical maximum number of unique combinations of a tensor.}
\label{fig:problem}
\end{figure*}

Fig.~\ref{fig:problem} explains what we perceive as the bottleneck in information propagation of BNNs. Here, ${\bf A}$ denotes the discrete input feature map to the binary convolution. Assume ${\bf W}$'s represent the set of all unique weight kernels one can imagine. Given a kernel size $k$, the number of input channels $C$, and a single output channel per kernel, the total number of unique ${\bf W}_i$'s, $\forall i \in [n]$ will then be $n = 2^{k^2 \times C}$. The output feature maps ${\bf D}_1$ to ${\bf D}_n$ are discrete finite-alphabet tensors. Note that $n$ in this case is smaller than the theoretical maximum number of unique activations $N = (k^2\times C)^{H\times W}$, given a specific input ${\bf A}$. Impacted by the activation design of the previous layers, proper design of ${\bf A}$ could potentially minimize the gap between $n$ and $N$. Applying \sign{} on ${\bf D}_1$ to ${\bf D}_n$ maps them to their binary counterparts ${\bf B}_1$ to ${\bf B}_n$. In theory, it is possible to have $n$ unique binary feature maps ${\bf B}_i$'s, $\forall i \in [n]$, even though in practice, we show the fact that if we give W the freedom to take any value, the set of possible tokens of the output feature map is limited for the \sign. Low diversity means that multiple distinct values of W will lead to an identical binary output, which will hinder the optimization of the model. We dub the aforementioned issue as the \emph{uniqueness bottleneck}. We argue that due to this bottleneck, the network does not utilize its full potential and the representational capacity of the network is going to be impacted. 

To demonstrate that this hypothesis is valid, we design a toy experiment that uses single-layer binary input feature maps ${\bf A}$ (in this case with $C = 1$) extracted from the binarized feature maps (using \sign{}) in a trained Bi-RealNet-$18$ \cite{liu2018bi}. Kernel size $k$ is set to $3$, which makes up for a total of $2^{k^2} = 512$  unique kernels. Following the steps sketched in Fig.~\ref{fig:problem}, we take ${\bf A}$ as the starting point, convolve it with every possible kernel ${\bf W}_i$ and binarize the output activations ${\bf D}_i$'s with the \sign{} function. We then count the number of unique binary feature maps ${\bf B}_i$'s, and average over all the channels (per different layers) of $20$ different input images. We denote the ratio of counted unique feature maps ($n_c$), and theoretical total number of unique feature maps ($n_t$) as the uniqueness ratio $\eta = n_c/n_t$. The results are shown in Table~\ref{tab:unique}. We can see that going deeper with convolutions, leading to smaller feature maps for layers $9$ to $16$, the uniqueness ratio decreases and the bottleneck becomes more evident. In the next section, we propose a learnable activation binarizer (\ourmethod) as a remedy for this bottleneck. 


\begin{table}[t]
\centering
\caption{$\eta$ for $512$ unique ${\bf W}$'s after applying \sign{} function in Bi-Real Net. Later layers show a lower ratio, indicating higher presence of the uniqueness bottleneck.}
\scriptsize
\setlength{\tabcolsep}{6pt}
\resizebox{\linewidth}{!}{
\label{tab:unique}
\begin{tabular}{@{}l|llllllll@{}}
\toprule
\textbf{Layer} & \multicolumn{1}{c}{\textbf{1}}  & \multicolumn{1}{c}{\textbf{2}}  & \multicolumn{1}{c}{\textbf{3}}  & \multicolumn{1}{c}{\textbf{4}}  & \multicolumn{1}{c}{\textbf{5}}  & \multicolumn{1}{c}{\textbf{6}}  & \multicolumn{1}{c}{\textbf{7}}  & \multicolumn{1}{c}{\textbf{8}}  \\
\textbf{$\eta$} & 0.964 & 0.994 & 0.996 & 0.998 & 0.998  & 0.986 & 0.991 & 0.994  \\ \midrule
\textbf{Layer} & \multicolumn{1}{c}{\textbf{9}} & \multicolumn{1}{c}{\textbf{10}} & \multicolumn{1}{c}{\textbf{11}} & \multicolumn{1}{c}{\textbf{12}} & \multicolumn{1}{c}{\textbf{13}} & \multicolumn{1}{c}{\textbf{14}} & \multicolumn{1}{c}{\textbf{15}} & \multicolumn{1}{c}{\textbf{16}} \\
\textbf{$\eta$} & 0.994 & 0.927 & 0.943  & 0.951 & 0.959 & 0.747 & 0.781 & 0.803                          \\ \bottomrule
\end{tabular}
}
\vspace{-2mm}
\end{table}

\section{The Proposed Method: \ourmethod}
\begin{figure}[t!]
\centering
    \includegraphics[width=\linewidth]{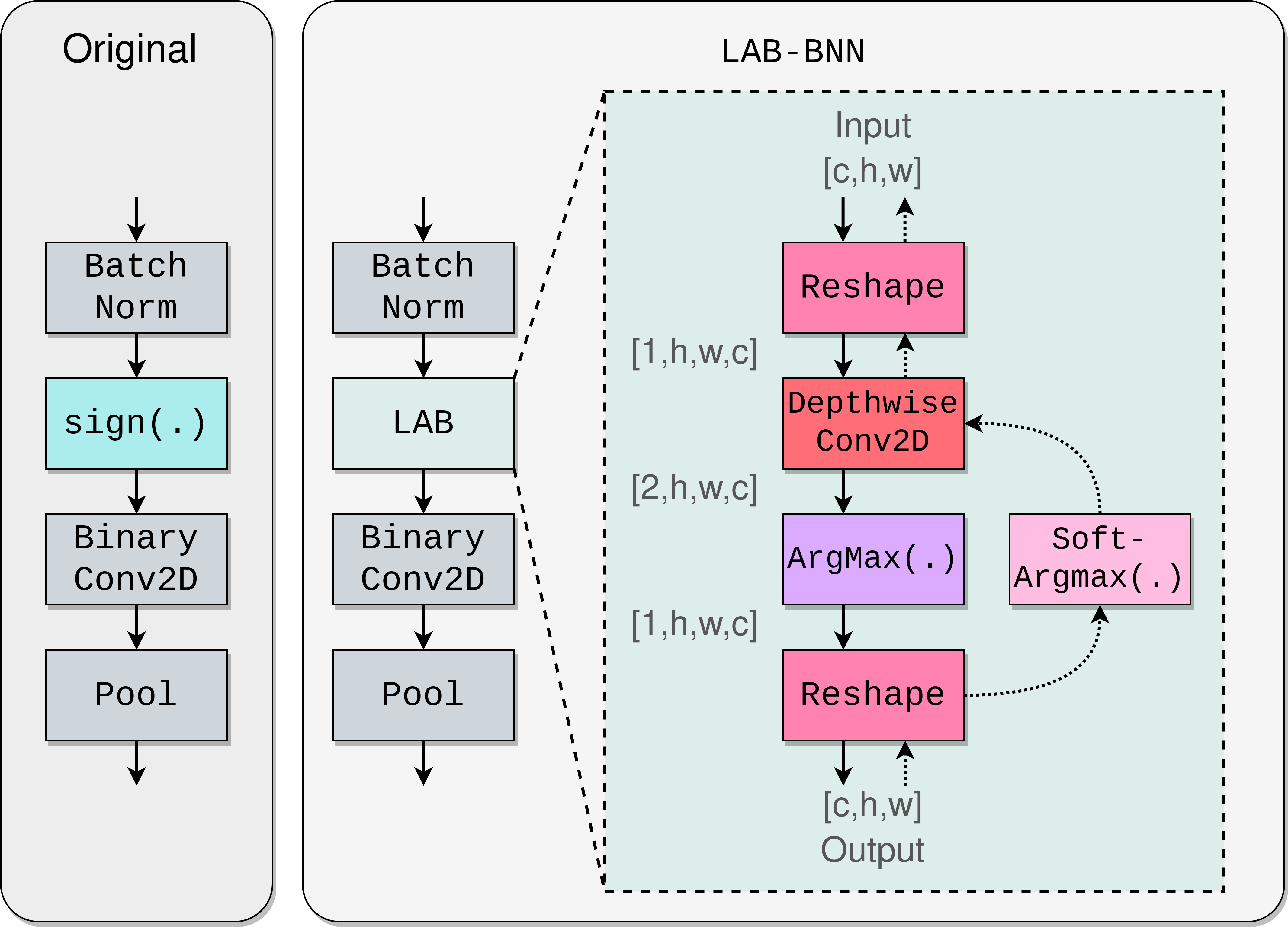}
  \captionof{figure}{Overview of \ourmethod{} and how it can be used similar to \sign. Tuples $[\{1, 2\}, H, W, C]$ indicate the shape of the tensors. The depthwise convolution together with \texttt{ArgMax(.)} form the core of \ourmethod. For differentiability in the backward pass , the \texttt{Soft-Argmax(.)} is used.}
  \label{fig:architecture}
  \vspace{-4mm}
\end{figure}

\vspace{-2mm}
One possible approach towards addressing the bottleneck of \sign{} binarization is to find a mapping from full-precision activation values to corresponding binary values, in such a way that the embedded spatial information from the input feature map is preserved. To do so, we propose to forge a different path in contrast to the current wisdom of activation distribution shaping \cite{ding2019regularizing, liu2020reactnet, kim2021improving}. More concretely, we propose a novel learnable activation binarizer (\ourmethod) to \emph{learn} a binarization kernel per layer, as shown in Fig.~\ref{fig:architecture}. The figure demonstrates \ourmethod{} as a building block of a standard BNN. Zooming into the \ourmethod{} unit, as we need to apply channel-wise binarization like \sign, the input is first reshaped for per-channel operations. To capture local spatial information per channel, a $3\times3$ depthwise convolution with a channel multiplier of $2$ is applied. The core idea behind this channel doubling is to construct a \emph{miniature segmentation layer} within the \ourmethod{} unit to classify the input as $-1$ or $+1$. This classification is done through an $\texttt{ArgMax(.)}$ across both channels, reducing the feature map back to a single output channel which is finally reshaped back to its original size. As the $\texttt{ArgMax(.)}$ is non-differentiable, we apply the $\texttt{Soft-ArgMax(.)}$ for the backward pass \cite{finn2016deep}. Given that we are dealing with only two classes, the $\texttt{Soft-ArgMax(.)}$ in \eqref{eq:soft_argmax} simplifies to a single entry of the $\texttt{SoftMax(.)}$ with an extra temperature controlling parameter $\beta$ which controls the ``hardness'' of the $\texttt{ArgMax(.)}$ approximation:
\vspace{-4mm}

\begin{equation}
\label{eq:soft_argmax}
\operatorname{Soft-ArgMax}(x)=\sum_{i=0}^{1} \frac{e^{\beta x_{i}}}{\sum_{j} e^{\beta x_{j}}} i = \frac{e^{\beta x_{1}}}{\sum_{j} e^{\beta x_{j}}}.
\end{equation}

The new binarization approach is only used for binarization of the feature maps and not for the binary convolution kernels, because the kernels do not contain enough spatial information for \ourmethod{} to capture.

\ourmethod{} ensures that the issues (ii), (iii) and (iv), introduced in section \ref{sec:intro} are solved. The learnable depthwise kernel ensures that the binary value of a pixel is dependent its neighbouring pixels. Because of this property, more diverse binary feature maps will be constructed, resulting in lifting of the uniqueness bottleneck. We realize that the introduction of a full-precision depthwise convolution adds additional complexity to the network. However, we argue that the network with \ourmethod{} is still a BNN, as the main convolutions are still binary; the depthwise convolution is an intermediate operation which is more often kept in full-precision in existing literature \cite{lin2021siman, lin2020rotated, liu2018rethinking, martinez2019training}.
\section{Experimental Evaluation}
In this section, we first reflect on the experimental setup. Next, we analyze the uniqueness bottleneck and how employing \ourmethod{} can help alleviate the problem. We then show that \ourmethod{} is beyond a one-off remedy but more of a universal module that can straightforwardly be plugged into existing baselines. Finally, we propose an end-to-end model architecture (\ournetwork) revolving around \ourmethod{} offering competitive performance against state-of-the-art. 

\subsection{Experimental Setup}
\label{sec:settings}
To be able to compare against a variety of existing state-of-the-art baselines - also to allow for a wider adoption by the community - we implement \ourmethod{} both in PyTorch and TensorFlow. For comparison with the state-of-the-art baselines, the TensorFlow-based Larq framework \cite{bannink2021larq} is used.

\textbf{Network structure.} To show the versatility of \ourmethod, we plug it into four different architectures. First, XNORNet \cite{rastegari2016xnor}, is chosen to examine the improved information flow on an AlexNet-based architecture \cite{krizhevsky2012imagenet} with no skip connections. Then, Bi-RealNet \cite{liu2018bi} and ReActNet \cite{liu2020reactnet} are used for their ResNet \cite{he2016deep} and MobileNet \cite{howard2017mobilenets} backbones, respectively. Finally, QuickNet \cite{bannink2021larq}, an improved version of Bi-RealNet, is used to assess whether \ourmethod{} can make an impact on a top performing state-of-the-art network. 

\textbf{Proposed end-to-end network: \ournetwork.} Going beyond the proposed module \ourmethod, we also design an end-to-end network based upon the architecture of Bi-RealNet-$18$\footnote{ReActNet-A was tried, but we could not reproduce in Larq.}, which is further enhanced by combining \texttt{PRelu} in \cite{tang2017train, bulat2019improved, liu2020reactnet, martinez2019training} and initial layers (further referred to as the \texttt{STEM}) as proposed by QuickNet \cite{bannink2021larq}.

\textbf{Hyperparameters.} All experiments are conducted using $4$ NVIDIA GeForce GTX $1080$ Ti GPUs and follow standard settings in Larq \cite{bannink2021larq}, unless otherwise mentioned. To reproduce nominal reported performance, we used a batch size of $128$ and a learning rate of $1e^{-4}$ for the retraining of XNORNet. For Bi-RealNet a batch size of $256$ and learning rate $2.5e^{-3}$ was used. For Quicknet we used a batch size of $512$ and a learning rate of $2.5e^{-3}$. Lastly, we re-trained ReActNet-A from scratch with a batch size of $128$ and learning rate of $2.5e^{-3}$. \ournetwork{} uses a batch size of $256$ and learning rate of $2.5e^{-3}$ and is trained from scratch for $300$ epochs. We chose to train from scratch as recent studies \cite{bethge2019back} suggest that multi-stage training is not needed for high accuracy models, and that it simplifies the training process significantly. The temperature controlling parameter $\beta$ of the $\texttt{Soft-ArgMax(.)}$ is made a learnable parameter, initialized with the value $1.0$.

\textbf{Real-time inference on the edge.} The research in BNNs is focussed on bringing deep learning to \emph{resource-constrained edge} devices. Recent studies report the computational complexity of their models using theoretical metrics such as floating-point operations (FLOPs) \cite{liu2020reactnet, martinez2019training} multiply-accumulate (MACs) \cite{bethge2019back} or arithmetic computation effort (ACE) \cite{zhang2021pokebnn}. In coherence with \cite{bannink2021larq, qin2021distribution} we argue that latency is the best metric to compare model performances. In order to benchmark model latency,
we use a resource-constrained edge computing device, Nvidia Jetson Xavier NX\footnote{https://www.nvidia.com/en-us/autonomous-machines/embedded-systems/jetson-xavier-nx/, accessed August $28$, $2022$} development kit, which is an ARM-based board for development of embedded AI systems. Although this device has a built-in GPU, for the benchmarking exercise we only use the CPU. Thus, these benchmarks can be reproduced on commodity ARM$64$ devices. To convert models from TensorFlow to Jetson-compatible models, we use the Larq Compute Engine Converter \cite{bannink2021larq}, which outputs a TensorFlow Lite (TFLite \cite{lite2019deploy}) model. This model can now be evaluated using a Larq benchmark tool \cite{bannink2021larq} adapted from the TFLite benchmark\footnote{https://www.tensorflow.org/lite/performance/measurement, accessed August $28$, $2022$}. For all the benchmarking experiments, the power mode of the Jetson is set to $15$W, we use the single thread mode, and report values averaged over $50$ runs.

\subsection{The Uniqueness Bottleneck: Qualitative and Quantitative Analysis}

\begin{figure}[t!]
\centering
\includegraphics[width=\linewidth]{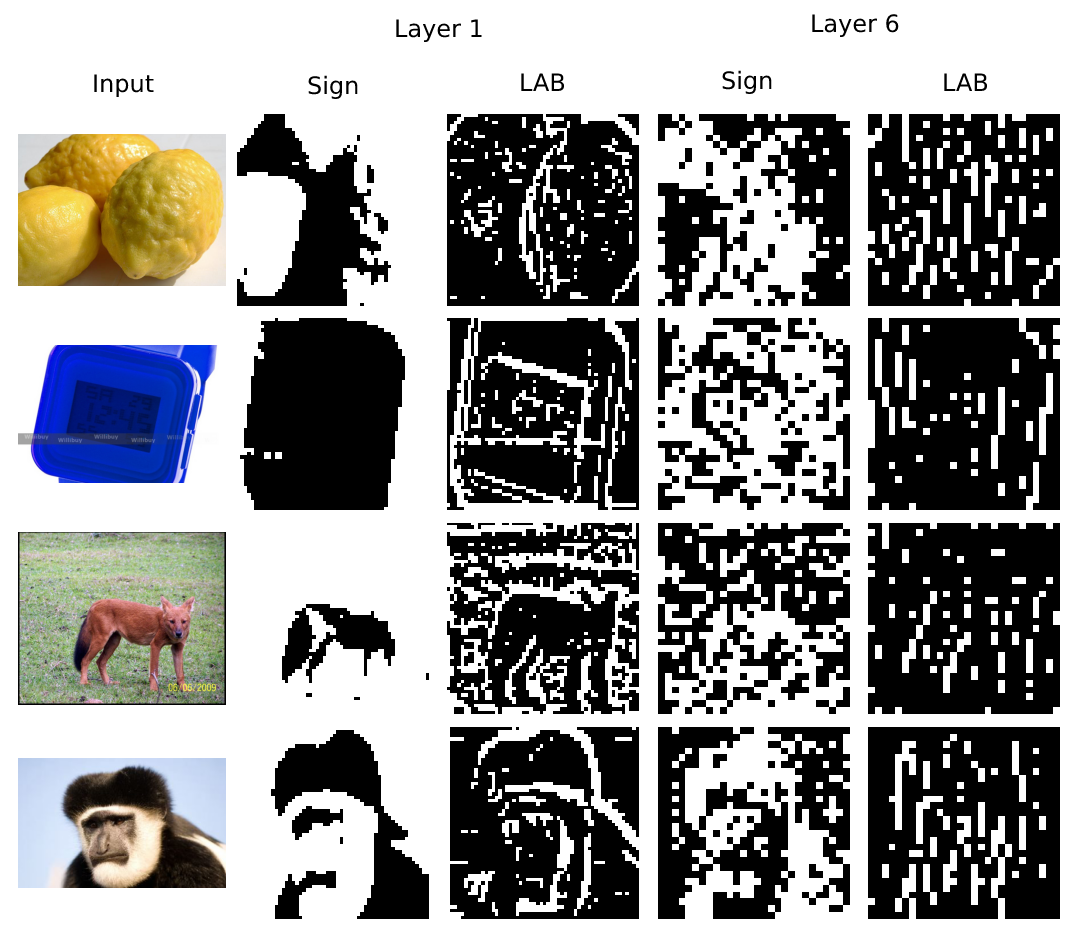}
\caption{Qualitative comparison between \sign{} and \ourmethod{} on two layers of Bi-RealNet-$18$. \ourmethod{} illustrates higher amount of texture (especially in Layer $1$), which indicates allowing more information to pass through.}
\label{fig:act_comparison}
\end{figure}

In section \ref{sec:problem}, we have shown that \sign{} can introduce a bottleneck in reaching the theoretical maximum number of unique binary states, which we argued would limit the capacity of BNNs. Here, we adopt a small-scale experiment (followed by large-scale end-to-end experiments in the next subsection) to qualitatively and quantitatively demonstrate that \ourmethod{} reduces this bottleneck. To this aim, we compare the binary feature maps of a trained Bi-RealNet-$18$ with a trained Bi-RealNet-$18+$\ourmethod{} (where \sign{} is replaced with \ourmethod). The results are shown in Fig.~\ref{fig:act_comparison} and Table~\ref{tab:ssim}. The figure depicts the selected feature maps from layer $1$ and $6$ (out of $18$ layers). As can be seen, more structural information is preserved in layer $1$ of \ourmethod{} compared to \sign, even though the features become too abstract for human understanding as we go to the deeper layer $6$. The most important takeaway from Fig. \ref{fig:act_comparison} is to understand that \ourmethod{}, in contrast to \sign{} can distinguish between similar pixel values (e.g. shades in the lemons and different tints of blue in the watch). Besides the qualitative demonstration, the impact of \ourmethod{} is further quantified on the number of unique feature maps learned by both networks through two (dis)similarity measures: structural similarity loss (SSIM) \cite{wang2004image} and a custom-designed metric we call Euclidean norm dissimilarity (\ourmetric{}), given by: 
\begin{equation}
    \label{eq:custom_metric}
    \resizebox{0.9\hsize}{!}{$%
    \begin{split}
            \ourmetric({\bf A}_i, {\bf A}_j) = \sqrt{\left(\frac{1}{HW}\sum_{w,h}^{}|{\bf A}_i-{\bf A}_j|\right)^2 + \left(\frac{1}{HW}\sum_{w,h}^{}|{\bf A}_i+{\bf A}_j|\right)^2}, \\ 
    h \in [H], w \in [W], i \neq j \in [C] 
    \end{split}
    $%
    }%
\end{equation}

where $W$ and $H$ denote the width and height of the input images in pixels and ${\bf A}_i$ and ${\bf A}_j$, are different single channel feature maps being compared. The metric has two terms in which the first term measures discrepancy, and the second one ensures fully inverted feature maps are not penalized. The latter is because inverted feature map layers along the channels can occur but do not indicate structural differences. Both measures are computed and averaged over all possible combinations of feature map layer pairs across all the $16$ convolutional layers of $10$ randomly selected images of ImageNet passed through both networks. The results are summarized in Table~\ref{tab:ssim}, and as can be seen, except for the first layer, \ourmethod{} shows higher dissimilarity values (a lower SSIM and a higher \ourmetric{}) indicating more uniqueness along the channels. 
\begin{table}[t!]
\setlength{\tabcolsep}{6pt}
\caption{Dissimilarity comparison between pre- and post-binarization of \sign{} vs. \ourmethod{} for SSIM and \ourmetric{} \eqref{eq:custom_metric} applied to extracted featuremaps from the trained BiRealNet-$18$ networks (with and without \ourmethod). The direction of the arrow indicates higher dissimilarity. The SSIM values are multiplied by $\times 10^3$.}
\centering
\resizebox{\linewidth}{!}{
\label{tab:ssim}
\begin{tabular}{lllccccccc}
\toprule 
\textbf{}                                             & \textbf{Layer} & \textbf{1} & \textbf{2}  & \textbf{3}  & \textbf{4}  & \textbf{5}  & \textbf{6}  & \textbf{7}  & \textbf{8}  \\ \midrule \midrule
\multicolumn{1}{l|}{\multirow{2}{*}{\textbf{SSIM$\downarrow$} ($\times 10^{-3}$)}}   & \textbf{Sign}  & 94.8     & 21.3      & 14.5      & 13.1      & 11.9      & 21.6      & 16.6      & 13.5      \\
\multicolumn{1}{l|}{}                                 & \textbf{LAB}   & 1.5     & 9.0      & 4.2      & 5.4      & 8.9      & 5.1      & 6.6      & 6.4      \\ \cline{2-10} 
\multicolumn{1}{l|}{\multirow{2}{*}{\textbf{\ourmetric$\uparrow$}}} & \textbf{Sign}  & 1.1        & 0.93        & 0.85        & 0.83        & 0.8         & 0.86        & 0.84        & 0.8         \\
\multicolumn{1}{l|}{}                                 & \textbf{LAB}   & 0.92       & 0.96        & 1.1         & 1.3         & 1.4         & 1.4         & 1.5         & 1.5         \\ \hline
                                                      & \textbf{Layer} & \textbf{9} & \textbf{10} & \textbf{11} & \textbf{12} & \textbf{13} & \textbf{14} & \textbf{15} & \textbf{16} \\ \hline
\multicolumn{1}{l|}{\multirow{2}{*}{\textbf{SSIM$\downarrow$} ($\times 10^{-3}$)}}   & \textbf{Sign}  & 12.2     & 25.8      & 19.7      & 17.6      & 14.6      & 29.3      & 22.4      & 18.0      \\
\multicolumn{1}{l|}{}                                 & \textbf{LAB}   & 8.5     & 4.2      & 2.9      & 2.4      & 8.8      & 5.6      & 8.2      & 7.1      \\ \cline{2-10} 
\multicolumn{1}{l|}{\multirow{2}{*}{\textbf{\ourmetric$\uparrow$}}} & \textbf{Sign}  & 0.76       & 0.92        & 0.88        & 0.85        & 0.81        & 0.98        & 0.93        & 0.89        \\
\multicolumn{1}{l|}{}                                 & \textbf{LAB}   & 1.6        & 1.4         & 1.4         & 1.4         & 1.7         & 1.4         & 1.4         & 1.4         \\ \hline 
\end{tabular}
}
\end{table}
%

\subsection{LAB: A Universal Module}
\vspace{-2mm}
\label{sec:universal}
The proposed method \ourmethod{} can readily be applied to any BNN, with no architectural adjustments. In this subsection, we demonstrate this by replacing the \sign{} with \ourmethod{} in four seminal BNN architectures and evaluate the impact on the downstream classification task of ImageNet. The results are illustrated in Table~\ref{tab:results_basic}. As can be seen, at the cost of negligible increase in model size and tolerable increase in delay (max of $2$x in [ms]), the boosted architectures XNORNet, Bi-RealNet, QuickNet, and ReActNet offer $\textbf{3.3\%}$, $\textbf{7.9\%}$, $\textbf{3.8\%}$ and $\textbf{1.7\%}$ improved Top-$1$ accuracies, respectively. The performance boost is slightly less pronounced for Top-$5$ accuracies. Note that the number of required training epochs to reach the nominal performance is dependent upon the architecture itself. The key message from this experiment is that irrespective of the backbone and architecture design, \ourmethod{} makes a considerable impact on all architectures.

\begin{table}[t]
    \centering
    \caption{Results of applying \ourmethod{} on the corresponding baselines on ImageNet.}
    \label{tab:results_basic}
    \setlength{\tabcolsep}{4pt}

    \resizebox{\linewidth}{!}{
\begin{tabular}{llcccccc}
\toprule
\multicolumn{1}{c}{\multirow{2}{*}{\textbf{Network}}}            & \multicolumn{1}{c}{\multirow{2}{*}{\textbf{Backbone}}}                & \multicolumn{1}{l}{\multirow{2}{*}{\textbf{Epochs}}} & \multirow{2}{*}{\textbf{Method}}     & \textbf{Top-1} & \textbf{Top-5}                 & \textbf{Model Size} & \textbf{Latency} \\
\multicolumn{1}{c}{}                                             &                                                   & \multicolumn{1}{l}{}                                 &                                      & [\%]           & [\%]                           & [MB]                & [ms]             \\ \midrule \midrule
\multicolumn{1}{l|}{\multirow{2}{*}{XNOR-Net \cite{rastegari2016xnor}}} & \multicolumn{1}{c|}{\multirow{2}{*}{AlexNet \cite{krizhevsky2012imagenet}}}     & \multicolumn{1}{c|}{\multirow{2}{*}{60}}             & \multicolumn{1}{c|}{Sign}            & 44.0           & \multicolumn{1}{c|}{68.1}      & 23.9                & 50.8             \\
\multicolumn{1}{l|}{}                                            & \multicolumn{1}{l|}{}                             & \multicolumn{1}{c|}{}                                & \multicolumn{1}{c|}{\mycc\ourmethod} & \mycc46.5      & \multicolumn{1}{c|}{\mycc70.3} & \mycc24.4           & \mycc55.1        \\ \hline
\multicolumn{1}{l|}{\multirow{2}{*}{BiRealNet \cite{liu2018bi}}}        & \multicolumn{1}{c|}{\multirow{2}{*}{ResNet-18 \cite{he2016deep}}}   & \multicolumn{1}{c|}{\multirow{2}{*}{150}}            & \multicolumn{1}{c|}{Sign}            & 54.4           & \multicolumn{1}{c|}{77.6}      & 4.18                & 72.5             \\
\multicolumn{1}{l|}{}                                            & \multicolumn{1}{l|}{}                             & \multicolumn{1}{c|}{}                                & \multicolumn{1}{c|}{\mycc\ourmethod} & \mycc59.1      & \multicolumn{1}{c|}{\mycc81.2} & \mycc4.65           & \mycc100.6       \\ \hline
\multicolumn{1}{l|}{\multirow{2}{*}{QuickNet \cite{bannink2021larq}}}   & \multicolumn{1}{c|}{\multirow{2}{*}{ResNet-18 \cite{he2016deep}}}   & \multicolumn{1}{c|}{\multirow{2}{*}{120}}            & \multicolumn{1}{c|}{Sign}            & 58.7           & \multicolumn{1}{c|}{81.2}      & 4.35                & 58.1             \\
\multicolumn{1}{l|}{}                                            & \multicolumn{1}{l|}{}                             & \multicolumn{1}{c|}{}                                & \multicolumn{1}{c|}{\mycc\ourmethod} & \mycc62.5      & \multicolumn{1}{c|}{\mycc84.0} & \mycc4.85           & \mycc82.4        \\ \hline
\multicolumn{1}{l|}{\multirow{2}{*}{ReActNet \cite{liu2020reactnet}}}   & \multicolumn{1}{c|}{\multirow{2}{*}{MobileNetV1 \cite{howard2017mobilenets}}} & \multicolumn{1}{c|}{\multirow{2}{*}{75}}             & \multicolumn{1}{c|}{Sign}            & 62.4           & \multicolumn{1}{c|}{83.4}      & 7.74                & 108.1            \\
\multicolumn{1}{l|}{}                                            & \multicolumn{1}{l|}{}                             & \multicolumn{1}{c|}{}                                & \multicolumn{1}{c|}{\mycc\ourmethod} & \mycc64.1      & \multicolumn{1}{c|}{\mycc84.8} & \mycc8.69           & \mycc210.9       \\ \bottomrule
\end{tabular}
    }
    \label{tab:results}
\vspace{-2mm}
\end{table}

\subsection{Comparison Against State-of-the-Art}
Now that the universality of \ourmethod{} has been shown, the efficacy of the proposed end-to-end network (\ournetwork) with \ourmethod{} sitting at its core, is evaluated. We compare the performance of \ournetwork{} against the state-of-the-art baselines focused on activation binarization. For fair comparison, we focus on reported Top-$1$ and $5$ validation accuracies on ImageNet with a ResNet-$18$ backbone. The outcome is summarized in Table~\ref{tab:results_sota}. Both Top-$1$ and $5$ results are competitive with the state-of-the-art and come short of the full precision equivalent network by only about $5\%$. Notably, ReActNet reports a higher accuracy (that we could not reproduce in TensorFlow) in part owing to adopting a multi-stage training strategy, whereas \ournetwork{} is trained from scratch. 
 
Aside from accuracy, the time and computational complexity of a BNN is just as important, even though most existing baselines overlook the importance of these metrics and sometimes do not even report them. Following \cite{liu2020reactnet}, in Table~\ref{tab:results_sota} we report the total binary and floating-point operations (OP $=$ BOP$/64$ $+$ FLOP). Here, we achieve a lower total computational complexity (in FLOPs) compared to other baselines including the original Bi-RealNet. To further break this down, in Table~\ref{tab:flop_breakdown} we set Bi-RealNet as our reference and state the added ($+$) complexity per operation listed in the second column. Even though the mechanics of \ourmethod{} adds slight complexity, we compensate for this with a more efficient implementation of the \texttt{STEM} layer \cite{bannink2021larq} leading to an overall decrease of $73.16 \times 10^{6}$ in total FLOPs.

\begin{table}[t]
    \centering
    \caption{Comparison against reported state-of-the-art on ImageNet. A dash indicates no value was reported by the original authors.}
    \label{tab:results_sota}
    
    \setlength{\tabcolsep}{4pt}

    \resizebox{\linewidth}{!}{

    \begin{tabular}{llllllll}
\toprule
\multicolumn{1}{c}{\multirow{2}{*}{\textbf{Network}}} & \multicolumn{1}{c}{\multirow{2}{*}{\textbf{Method}}} & \multicolumn{1}{c}{\multirow{2}{*}{\textbf{W/A}}} & \multicolumn{1}{c}{\textbf{Top-1}} & \multicolumn{1}{c}{\textbf{Top-5}} & \multicolumn{1}{c}{\textbf{BOPs}} & \multicolumn{1}{c}{\textbf{FLOPs}} & \multicolumn{1}{c}{\textbf{OPs}} \\
\multicolumn{1}{c}{}                                  & \multicolumn{1}{c}{}                                 & \multicolumn{1}{c}{}                              & \multicolumn{1}{c}{[\%]}                               & \multicolumn{1}{c}{[\%]}                               & \multicolumn{1}{c}{$(\times10^{9})$}                  & \multicolumn{1}{c}{$(\times10^{8})$}           & \multicolumn{1}{c}{$(\times10^{8})$}        \\ \midrule \midrule
\multirow{9}{*}{ResNet-18}                           & Full-precision                                       & \multicolumn{1}{|c|}{32/32}                                             & \multicolumn{1}{c}{69.6}                               & \multicolumn{1}{c|}{89.2}                               & \multicolumn{1}{c}{0}                                 & \multicolumn{1}{c}{18.1}                       & \multicolumn{1}{c}{18.1}                    \\ \cline{2-3} 
                                                      & Bi-RealNet \cite{liu2018bi}                          & \multicolumn{1}{|c|}{1/1}                                               & \multicolumn{1}{c}{56.4}                               & \multicolumn{1}{c|}{79.5}                               & \multicolumn{1}{c}{1.68}                              & \multicolumn{1}{c}{1.39}                       & \multicolumn{1}{c}{1.63}                    \\ \cline{2-3} 
                                                      & BNN-UAD \cite{kim2021improving}                      & \multicolumn{1}{|c|}{1/1}                                               & \multicolumn{1}{c}{57.2}                               & \multicolumn{1}{c|}{80.2}                               & \multicolumn{1}{c}{-}                                 & \multicolumn{1}{c}{-}                          & \multicolumn{1}{c}{-}                       \\ \cline{2-3} 
                                                      & IR-Net \cite{qin2020forward}                         & \multicolumn{1}{|c|}{1/1}                                               & \multicolumn{1}{c}{58.1}                               & \multicolumn{1}{c|}{80.0}                               & \multicolumn{1}{c}{-}                                 & \multicolumn{1}{c}{-}                          & \multicolumn{1}{c}{1.63}                    \\ \cline{2-3} 
                                                      & SI-BNN \cite{wang2020sparsity}                       & \multicolumn{1}{|c|}{1/1}                                               & \multicolumn{1}{c}{59.7}                               & \multicolumn{1}{c|}{81.8}                               & \multicolumn{1}{c}{-}                                 & \multicolumn{1}{c}{-}                          & \multicolumn{1}{c}{-}                       \\ \cline{2-3} 
                                                      & SiMaN \cite{lin2021siman}                            & \multicolumn{1}{|c|}{1/1}                                               & \multicolumn{1}{c}{60.1}                               & \multicolumn{1}{c|}{82.3}                               & \multicolumn{1}{c}{-}                                 & \multicolumn{1}{c}{-}                          & \multicolumn{1}{c}{-}                       \\ \cline{2-3} 
                                                      & QuickNet \cite{bannink2021larq}                      & \multicolumn{1}{|c|}{1/1}                                               & \multicolumn{1}{c}{63.3}                               & \multicolumn{1}{c|}{84.6}                               & \multicolumn{1}{c}{-}                                 & \multicolumn{1}{c}{-}                          & \multicolumn{1}{c}{-}                       \\ \cline{2-3} 
                                                      & \mycc\ournetwork                                     & \multicolumn{1}{|c|}{\mycc1/1}                                          & \multicolumn{1}{c}{\mycc64.2}                          & \multicolumn{1}{c|}{\mycc85.0}                          & \multicolumn{1}{c}{\mycc1.68}                         & \multicolumn{1}{c}{\mycc0.66}                  & \multicolumn{1}{c}{\mycc0.92}                    \\ \cline{2-3} 
                                                      & ReActNet \cite{liu2020reactnet}                      & \multicolumn{1}{|c|}{1/1}                                               & \multicolumn{1}{c}{65.9}                               & \multicolumn{1}{c|}{-}                                  & \multicolumn{1}{c}{-}                                 & \multicolumn{1}{c}{-}                          & \multicolumn{1}{c}{-}                       \\  \bottomrule
\end{tabular}
}
\end{table}
\begin{table}[t]
    \centering
    \caption{FLOPs comparison: \ournetwork{} vs.  Bi-RealNet.}
    \setlength{\tabcolsep}{8pt}
    \resizebox{0.65\linewidth}{!}{    \label{tab:flop_breakdown}

        \begin{tabular}{lcc}
        \toprule
        \multicolumn{1}{l}{\multirow{2}{*}{\textbf{Component}}} & \multirow{2}{*}{\textbf{Operation}} & \textbf{FLOPs}                                                  \\
        \multicolumn{1}{c}{}                                 &                                    & \multicolumn{1}{l}{$\displaystyle \left( \times 10^{6}\right)$} \\ \midrule \midrule
                                                              & \multicolumn{1}{|c|}{Depthwise Conv2D}                   & +30.3                                                           \\
                                                              & \multicolumn{1}{|c|}{BiasAdd}                            & +1.68                                                           \\
        \ourmethod{} (ours)                                              & \multicolumn{1}{|c|}{Multiply}                           & +1.68                                                           \\
                                                              & \multicolumn{1}{|c|}{ArgMax}                             & +0.84                                                           \\
                                                              & \multicolumn{1}{|c|}{Equal}                              & +0.84                                                           \\ \hline
        \texttt{PRelu} \cite{tang2017train}                                                 & \multicolumn{1}{|c|}{Mul}                                & +0.57                                                           \\
                                                              & \multicolumn{1}{|c|}{Neg}                                & +0.76                                                           \\ \hline
        STEM \cite{bannink2021larq}                                         & \multicolumn{1}{|c|}{Conv2D}                             & -109.83                                                         \\ \midrule
        \textbf{Total}                   &                                    & \textbf{-73.16}                                                 \\ \bottomrule
        \end{tabular}
    }
    
\vspace{-6mm}
\end{table}

\begin{figure}[h]
    \centering
    \includegraphics[trim={10px 0 15px 0},clip, width=\linewidth]{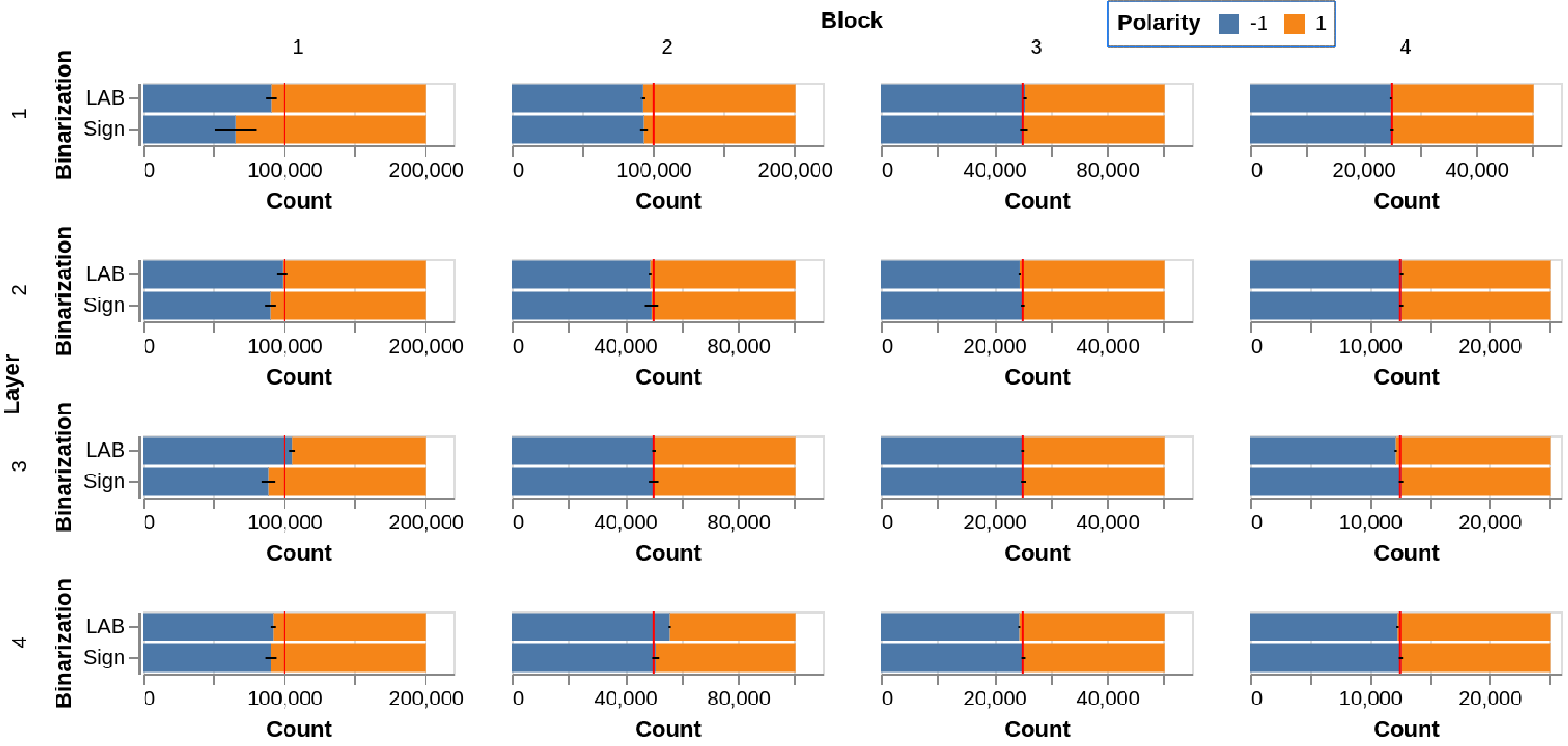}
    \caption{Post-binarization distribution of original Bi-RealNet-$18$ vs \ournetwork{} averaged over all channels of $1000$ input images. Rows indicate the blocks in the ResNet structure, columns indicate layer number within each block.}
    \label{fig:act_dist}
    \vspace{-1mm}
\end{figure}

\textbf{Post-binarization distribution.} To illustrate that the post-binarization distribution is not a telling factor about the performance of the network (in contrast to conclusions drawn in \cite{ding2019regularizing, kim2021improving, liu2020reactnet}), Fig.~\ref{fig:act_dist} shows the post-binarization distribution of $+1$'s and $-1$'s for original Bi-RealNet-$18$ and Bi-RealNet-$18+$\ourmethod{} averaged over all channels of $1000$ input images. As can be seen, while the distribution of binary values remains almost the same across both networks, the performance of the two (reported in Subsection~\ref{sec:universal}) is apart by $7.9\%$, supporting our claim that the binary distribution is not important, but the spatial arrangement of pixels is.

\textbf{Going deep or going \ourmethod?} Plugging \ourmethod{} into the standard Bi-RealNet (our base to build \ournetwork) adds additional complexity to the network, and one could argue that adding the kernel of \ourmethod{} per layer could virtually help make the network deeper. However, as a counter-argument, we compare \ournetwork{} against a Bi-RealNet with a deeper backbone. More specifically, in the current construct, our method adds a depthwise convolution for each binary layer in Bi-RealNet-$18$. This is (roughly) equivalent to addition of $16$ convolution layers, yielding a total of $18 + 16 = 34$ layers. According to \cite{liu2018bi}, Bi-RealNet-$34$ and Bi-RealNet-$50$ respectively achieve $62.2\%$ and $62.6\%$ validation accuracies on ImageNet which are both still below the accuracy of \ournetwork{}. Furthermore, the model size of BiRealNet-18+\ourmethod{} is $4.24$MB (FP32 weights for depthwise kernel), whereas BiRealNet-34 occupies $5.23$MB. Thus, we are offering smaller model size and higher accuracy. To assess the impact of precision on the accuracy gain of \ourmethod, we evaluated the depthwise convolution kernel of \ourmethod{} for different bit widths. The results summarized in Table \ref{tab:quantization} indicate that going down from FP32 to INT8, the impact of \ourmethod{} does not degrade, and even INT4 still offers a significant gain over no-\ourmethod{} baseline. 

\textbf{Comparison against local thresholding} We compare \ourmethod{} against classical local binarization techniques, to show that \ourmethod{} adheres to the common wisdom that learnable approaches outperform statistical approaches. We implement Niblack \cite{niblack1985introduction} and Sauvola \cite{sauvola2000adaptive} and use a window size of $3$ and common values for $k$. Table \ref{tab:adatresh} shows \ourmethod{} has a gap of $11.5\%$ with Niblack and improves with $2.7\%$ over the best configuration for Sauvola.

\begin{table}[t]
\centering
\caption{\ournetwork{} (Imagenette, $100$ epochs) with multiple precisions for the depthwise convolution for \ourmethod, and a comparison with other adaptive binarization techniques.}
\label{tab:quantization}
\resizebox{0.75\linewidth}{!}{
\begin{tabular}{@{}cccc@{}}
\toprule
\ourmethod{} depthwise                 & \textbf{Top 1}            & \textbf{Top 5}            & \textbf{Model Size} \\
convolution                            & [\%]                      & [\%]                      & [MB]                \\ \hline \midrule
\multicolumn{1}{c|}{No-\ourmethod}     & \multicolumn{1}{c|}{76.4} & \multicolumn{1}{c|}{97.2} & 2.04                \\ \hline
\multicolumn{1}{c|}{\ourmethod{} FP32} & \multicolumn{1}{c|}{88.5} & \multicolumn{1}{c|}{99.2} & 2.30                \\ \hline
\multicolumn{1}{c|}{\ourmethod{} INT8} & \multicolumn{1}{c|}{89.1} & \multicolumn{1}{c|}{99.2} & 2.13                \\ \hline
\multicolumn{1}{c|}{\ourmethod{} INT4} & \multicolumn{1}{c|}{85.3} & \multicolumn{1}{c|}{98.7} & 2.10                \\ \bottomrule
\end{tabular}
}
\vspace{-2mm}
\end{table}

\begin{table}[t]
    \centering
    \captionsetup{justification=centering}
    \caption{\ournetwork{} (Imagenette, $150$ epochs, batch size $32$) A comparison with adaptive binarization techniques.} 
    \label{tab:adatresh}
\resizebox{0.8\linewidth}{!}{
\begin{tabular}{@{}cccc@{}}
\toprule
Binarization                           & \textbf{Top 1}            & \textbf{Top 5}            & \textbf{Model Size} \\
method                                 & [\%]                      & [\%]                      & [MB]                \\ \hline \midrule
\multicolumn{1}{c|}{STE-sign}          & \multicolumn{1}{c|}{87.0}     & \multicolumn{1}{c|}{98.7} & 2.06                 \\ \hline
\multicolumn{1}{c|}{Niblack, $k=-0.2$} & \multicolumn{1}{c|}{77.8}     & \multicolumn{1}{c|}{97.0} & 2.06                 \\ \hline
\multicolumn{1}{c|}{Sauvola, $k=0.2$}  & \multicolumn{1}{c|}{82.5}     & \multicolumn{1}{c|}{98.3} & 2.06                \\ \hline
\multicolumn{1}{c|}{Sauvola, $k=0.5$}  & \multicolumn{1}{c|}{86.6}     & \multicolumn{1}{c|}{98.8} & 2.06                \\ \hline
\multicolumn{1}{c|}{\ourmethod}        & \multicolumn{1}{c|}{89.3}     & \multicolumn{1}{c|}{99.1} & 2.09                \\ \bottomrule
\end{tabular}
}
\vspace{-4mm}
\end{table}

\vspace{-3mm}
\section{Ablation Studies}
\vspace{-2mm}
In this section, the effect of different components of \ournetwork{} is further inspected. In what follows, we use the same settings listed in Section~\ref{sec:settings}, except for the number of epochs which are shortened to $30$. We start with the original Bi-RealNet-$18$ (model \textbf{A}) and update the components progressively towards \ournetwork{} (model \textbf{D}). We first upgrade model \textbf{A} by incorporating \texttt{PRelu} activation \cite{tang2017train} resulting in \textbf{B}. Model \textbf{B} is then extended by replacing \sign{} by the proposed \ourmethod{} module leading to model \textbf{C}. Lastly, we replace the initial convolution in Bi-RealNet with the \texttt{STEM} layer of QuickNet \cite{bannink2021larq}, which forms \ournetwork{} (model \textbf{D}). The results are illustrated in Table~\ref{tab:ablation}. Plugging \ourmethod{} into model \textbf{B} results in an improvement of about $6\%$ on the Top-$1$ validation accuracy (and roughly $5\%$ for Top-$5$), at an increase of $0.46\text{MB}$ in model size and $28.5\text{ms}$ in latency. To make up for the added latency, we apply the \texttt{STEM} layers as proposed by QuickNet \cite{bannink2021larq} which results in decreasing the latency with respect to model \textbf{A} down to $9.7\text{ms}$.

\begin{figure*}[t!]
    \centering
    \includegraphics[width=.75\linewidth]{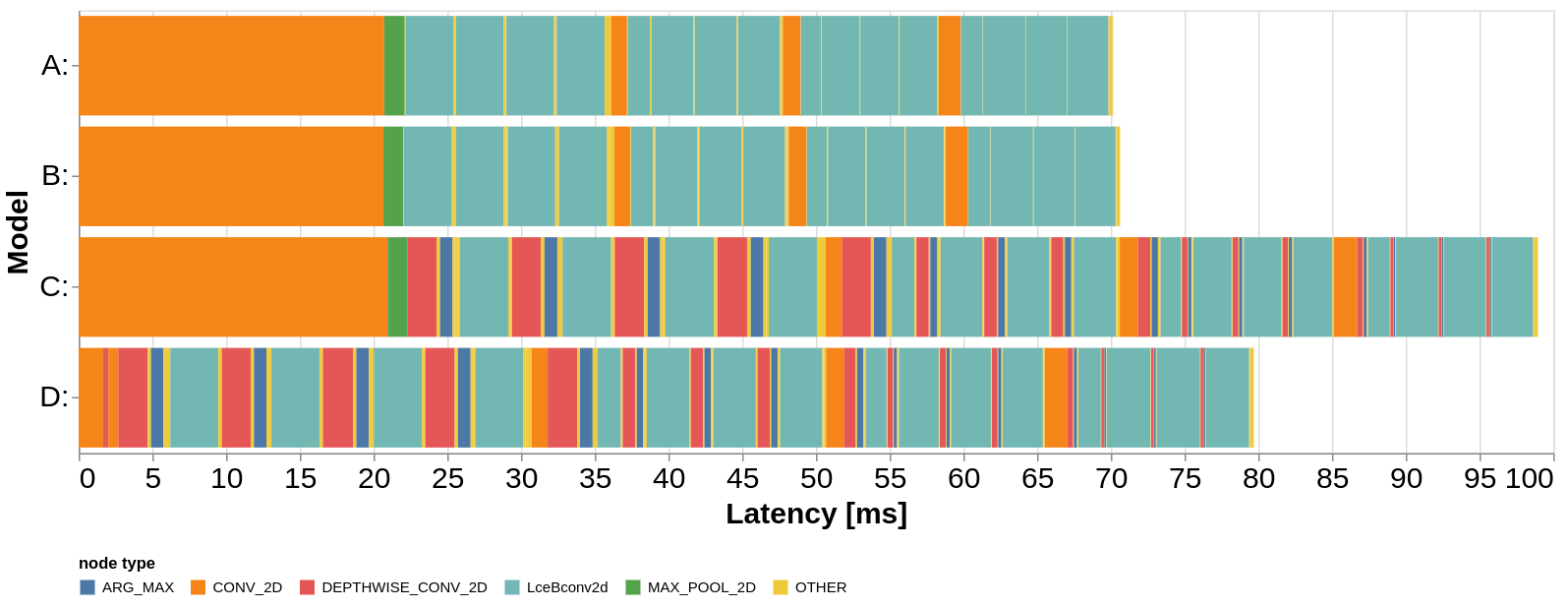}
    \caption{Breakdown of per-operator latencies for ablation study of \ournetwork. Operators contributing to the latency minimally have been marked as ``other''.}
    \label{fig:operator_profiling}
\end{figure*}

To provide further insights into the distribution of per-operation latencies, in Fig.~\ref{fig:operator_profiling} we profile the network delays for models \textbf{A} to \textbf{D}. In other words, this is a fine-grained visual breakdown of the total latencies reported in Table~\ref{tab:ablation}. The depthwise convolution together with the \texttt{ArgMax(.)} operation in \ourmethod{} are the main culprits behind the added latency. Additionally, it is clear that the \texttt{STEM} layer indeed helps to alleviate the overall latency as discussed in Table~\ref{tab:ablation}. 

\subsection{Where to apply \ourmethod.} 
So far, when applying \ourmethod, all instances of \sign{} in every binary layer of the network were replaced with \ourmethod. For the architecture Bi-RealNet-$18$, this means for all of the four residual blocks, in all four convolutions per block. As we saw in Table~\ref{tab:unique}, the uniqueness bottleneck is mainly visible in later layers. Therefore, in Table~\ref{tab:ablations_which_block} we assess in which combination of blocks (out of $16$ possible combinations) it is most impactful to apply \ourmethod{} as a binarization function. In doing so, we apply \ourmethod{} on all layers per selected block. The networks are trained for $30$ epochs and we report Top-$1$ and Top-$5$ validation accuracies, model size and and latencies. We conclude from the table that omitting \ourmethod{} (using the original \sign{} everywhere) leads to the worst results, and see that as we apply \ourmethod{} in more blocks progressively, the accuracy increases, and applying \ourmethod{} to every block gives the highest accuracy. Interestingly, the model with \ourmethod{} applied to the last 3 blocks (in the second row) has an exceptional accuracy-latency trade-off. Table \ref{tab:ablations_which_block} shows that by applying \ourmethod{} in different blocks across the network, it is possible to design a network with a certain accuracy-latency trade-off, which gives useful design freedom for the practical use cases of BNN's.

\begin{table}[t]
    \centering
    \caption{Ablation study for \ournetwork{}, trained for $30$ epochs on ImageNet dataset. The per-operator breakdown of the latency column is shown in Fig.~\ref{fig:operator_profiling}.} 
    \label{tab:ablation}
    \setlength{\tabcolsep}{3pt}
    \resizebox{\linewidth}{!}{ 
\begin{tabular}{@{}lcccc@{}}
\toprule
\multirow{2}{*}{\textbf{Method}}   & \textbf{Top-1} & \textbf{Top-5} & \textbf{Model Size} & \textbf{Latency} \\
                                   & [\%]           & [\%]           & [MB]                & [ms]             \\ \midrule \midrule
\textbf{A:} BiRealNet [Base]           & 45.0           & 69.7           & 4.18                & 70.3             \\
\textbf{B:} Base+\texttt{PRelu} & 52.4           & 76.1           & 4.19                & 70.8             \\
\textbf{C:} Base+\texttt{PRelu}+\ourmethod       & 58.1           & 80.9           & 4.65                & 99.3             \\
\textbf{D:} \textbf{C}+\texttt{STEM} [\ournetwork{}]  & 58.2           & 80.8           & 4.62                & 80.0             \\ \bottomrule
\end{tabular}
    }
\end{table}

\section{Concluding Remarks}
\vspace{-2mm}
We have shown that the commonly adopted binarization operation \sign{} imposes a \emph{uniqueness bottleneck} on BNNs, making it a sub-optimal choice for binarization. As a remedy, we introduce learnable activation binarizer (\ourmethod), a novel binarization function that allows BNNs to learn a flexible binarization kernel per layer. We have demonstrated that \ourmethod{} can readily be plugged into existing baseline BNNs boosting their performance regardless of their architecture design. Beyond that, we have also built a new end-to-end network (\ournetwork) based upon \ourmethod{} that offers competitive performance on par with the state-of-the-art. For future work, we will investigate applying learnable binarization to weights in addition to activations. We also plan to further extend our experimentation especially around \ournetwork{} to push the performance boundaries and advance the state-of-the-art.

\begin{table}[t]
        \centering
        \caption{Study on applying \ourmethod{} to different blocks of \ournetwork. A checkmark indicates that \ourmethod{} was used for all layers in that block.}  
        \label{tab:ablations_which_block} 
        
        \setlength{\tabcolsep}{3pt}
            \resizebox{0.7\linewidth}{!}{

\begin{tabular}{llllcccc}
\toprule
\multicolumn{4}{c}{\textbf{Block}}                                                                                                                                                                                                                                                                                                                                                        & \textbf{Top-1} & \textbf{Top-5} & \textbf{Model Size} & \textbf{Latency} \\
\multicolumn{1}{c}{\textbf{1}}                                                               & \multicolumn{1}{c}{\textbf{2}}                                                               & \multicolumn{1}{c}{\textbf{3}}                                                               & \multicolumn{1}{c}{\textbf{4}}                                                               & [\%]           & [\%]           & [MB]                & [ms]             \\ \midrule \midrule
\multicolumn{1}{l|}{\makebox[0pt][l]{$\square$}\raisebox{.15ex}{\hspace{0.1em}$\checkmark$}} & \multicolumn{1}{l|}{\makebox[0pt][l]{$\square$}\raisebox{.15ex}{\hspace{0.1em}$\checkmark$}} & \multicolumn{1}{l|}{\makebox[0pt][l]{$\square$}\raisebox{.15ex}{\hspace{0.1em}$\checkmark$}} & \multicolumn{1}{l|}{\makebox[0pt][l]{$\square$}\raisebox{.15ex}{\hspace{0.1em}$\checkmark$}} & 59.1           & 81.3           & 4.68                & 83.7             \\ \hline
\multicolumn{1}{l|}{\makebox[0pt][l]{$\square$}\raisebox{.15ex}{\hspace{1em}}}               & \multicolumn{1}{l|}{\makebox[0pt][l]{$\square$}\raisebox{.15ex}{\hspace{0.1em}$\checkmark$}} & \multicolumn{1}{l|}{\makebox[0pt][l]{$\square$}\raisebox{.15ex}{\hspace{0.1em}$\checkmark$}} & \multicolumn{1}{l|}{\makebox[0pt][l]{$\square$}\raisebox{.15ex}{\hspace{0.1em}$\checkmark$}} & 58.7           & 81.0           & 4.66                & 68.0             \\ \hline
\multicolumn{1}{l|}{\makebox[0pt][l]{$\square$}\raisebox{.15ex}{\hspace{0.1em}$\checkmark$}} & \multicolumn{1}{l|}{\makebox[0pt][l]{$\square$}\raisebox{.15ex}{\hspace{1em}}}               & \multicolumn{1}{l|}{\makebox[0pt][l]{$\square$}\raisebox{.15ex}{\hspace{0.1em}$\checkmark$}} & \multicolumn{1}{l|}{\makebox[0pt][l]{$\square$}\raisebox{.15ex}{\hspace{0.1em}$\checkmark$}} & 58.4           & 80.6           & 4.64                & 75.3             \\ \hline
\multicolumn{1}{l|}{\makebox[0pt][l]{$\square$}\raisebox{.15ex}{\hspace{0.1em}$\checkmark$}} & \multicolumn{1}{l|}{\makebox[0pt][l]{$\square$}\raisebox{.15ex}{\hspace{0.1em}$\checkmark$}} & \multicolumn{1}{l|}{\makebox[0pt][l]{$\square$}\raisebox{.15ex}{\hspace{0.1em}$\checkmark$}} & \multicolumn{1}{l|}{\makebox[0pt][l]{$\square$}\raisebox{.15ex}{\hspace{1em}}}               & 58.4           & 80.5           & 4.33                & 79.1             \\ \hline
\multicolumn{1}{l|}{\makebox[0pt][l]{$\square$}\raisebox{.15ex}{\hspace{0.1em}$\checkmark$}} & \multicolumn{1}{l|}{\makebox[0pt][l]{$\square$}\raisebox{.15ex}{\hspace{0.1em}$\checkmark$}} & \multicolumn{1}{l|}{\makebox[0pt][l]{$\square$}\raisebox{.15ex}{\hspace{1em}}}               & \multicolumn{1}{l|}{\makebox[0pt][l]{$\square$}\raisebox{.15ex}{\hspace{0.1em}$\checkmark$}} & 58.1           & 80.4           & 4.61                & 77.1             \\ \hline
\multicolumn{1}{l|}{\makebox[0pt][l]{$\square$}\raisebox{.15ex}{\hspace{1em}}}               & \multicolumn{1}{l|}{\makebox[0pt][l]{$\square$}\raisebox{.15ex}{\hspace{0.1em}$\checkmark$}} & \multicolumn{1}{l|}{\makebox[0pt][l]{$\square$}\raisebox{.15ex}{\hspace{0.1em}$\checkmark$}} & \multicolumn{1}{l|}{\makebox[0pt][l]{$\square$}\raisebox{.15ex}{\hspace{1em}}}               & 58.0           & 80.2           & 4.31                & 63.9             \\ \hline
\multicolumn{1}{l|}{\makebox[0pt][l]{$\square$}\raisebox{.15ex}{\hspace{1em}}}               & \multicolumn{1}{l|}{\makebox[0pt][l]{$\square$}\raisebox{.15ex}{\hspace{0.1em}$\checkmark$}} & \multicolumn{1}{l|}{\makebox[0pt][l]{$\square$}\raisebox{.15ex}{\hspace{1em}}}               & \multicolumn{1}{l|}{\makebox[0pt][l]{$\square$}\raisebox{.15ex}{\hspace{0.1em}$\checkmark$}} & 57.9           & 80.1           & 4.58                & 64.9             \\ \hline
\multicolumn{1}{l|}{\makebox[0pt][l]{$\square$}\raisebox{.15ex}{\hspace{1em}}}               & \multicolumn{1}{l|}{\makebox[0pt][l]{$\square$}\raisebox{.15ex}{\hspace{1em}}}               & \multicolumn{1}{l|}{\makebox[0pt][l]{$\square$}\raisebox{.15ex}{\hspace{0.1em}$\checkmark$}} & \multicolumn{1}{l|}{\makebox[0pt][l]{$\square$}\raisebox{.15ex}{\hspace{0.1em}$\checkmark$}} & 57.8           & 80.0           & 4.62                & 60.6             \\ \hline
\multicolumn{1}{l|}{\makebox[0pt][l]{$\square$}\raisebox{.15ex}{\hspace{0.1em}$\checkmark$}} & \multicolumn{1}{l|}{\makebox[0pt][l]{$\square$}\raisebox{.15ex}{\hspace{1em}}}               & \multicolumn{1}{l|}{\makebox[0pt][l]{$\square$}\raisebox{.15ex}{\hspace{0.1em}$\checkmark$}} & \multicolumn{1}{l|}{\makebox[0pt][l]{$\square$}\raisebox{.15ex}{\hspace{1em}}}               & 57.8           & 79.9           & 4.29                & 72.9             \\ \hline
\multicolumn{1}{l|}{\makebox[0pt][l]{$\square$}\raisebox{.15ex}{\hspace{1em}}}               & \multicolumn{1}{l|}{\makebox[0pt][l]{$\square$}\raisebox{.15ex}{\hspace{1em}}}               & \multicolumn{1}{l|}{\makebox[0pt][l]{$\square$}\raisebox{.15ex}{\hspace{0.1em}$\checkmark$}} & \multicolumn{1}{l|}{\makebox[0pt][l]{$\square$}\raisebox{.15ex}{\hspace{1em}}}               & 56.7           & 78.9           & 4.27                & 58.8             \\ \hline
\multicolumn{1}{l|}{\makebox[0pt][l]{$\square$}\raisebox{.15ex}{\hspace{0.1em}$\checkmark$}} & \multicolumn{1}{l|}{\makebox[0pt][l]{$\square$}\raisebox{.15ex}{\hspace{1em}}}               & \multicolumn{1}{l|}{\makebox[0pt][l]{$\square$}\raisebox{.15ex}{\hspace{1em}}}               & \multicolumn{1}{l|}{\makebox[0pt][l]{$\square$}\raisebox{.15ex}{\hspace{0.1em}$\checkmark$}} & 56.7           & 79.3           & 4.57                & 71.1             \\ \hline
\multicolumn{1}{l|}{\makebox[0pt][l]{$\square$}\raisebox{.15ex}{\hspace{0.1em}$\checkmark$}} & \multicolumn{1}{l|}{\makebox[0pt][l]{$\square$}\raisebox{.15ex}{\hspace{0.1em}$\checkmark$}} & \multicolumn{1}{l|}{\makebox[0pt][l]{$\square$}\raisebox{.15ex}{\hspace{1em}}}               & \multicolumn{1}{l|}{\makebox[0pt][l]{$\square$}\raisebox{.15ex}{\hspace{1em}}}               & 56.4           & 79.1           & 4.26                & 75.0             \\ \hline
\multicolumn{1}{l|}{\makebox[0pt][l]{$\square$}\raisebox{.15ex}{\hspace{1em}}}               & \multicolumn{1}{l|}{\makebox[0pt][l]{$\square$}\raisebox{.15ex}{\hspace{1em}}}               & \multicolumn{1}{l|}{\makebox[0pt][l]{$\square$}\raisebox{.15ex}{\hspace{1em}}}               & \multicolumn{1}{l|}{\makebox[0pt][l]{$\square$}\raisebox{.15ex}{\hspace{0.1em}$\checkmark$}} & 55.9           & 78.6           & 4.54                & 41.4             \\ \hline
\multicolumn{1}{l|}{\makebox[0pt][l]{$\square$}\raisebox{.15ex}{\hspace{1em}}}               & \multicolumn{1}{l|}{\makebox[0pt][l]{$\square$}\raisebox{.15ex}{\hspace{0.1em}$\checkmark$}} & \multicolumn{1}{l|}{\makebox[0pt][l]{$\square$}\raisebox{.15ex}{\hspace{1em}}}               & \multicolumn{1}{l|}{\makebox[0pt][l]{$\square$}\raisebox{.15ex}{\hspace{1em}}}               & 55.7           & 78.1           & 4.23                & 47.6             \\ \hline
\multicolumn{1}{l|}{\makebox[0pt][l]{$\square$}\raisebox{.15ex}{\hspace{0.1em}$\checkmark$}} & \multicolumn{1}{l|}{\makebox[0pt][l]{$\square$}\raisebox{.15ex}{\hspace{1em}}}               & \multicolumn{1}{l|}{\makebox[0pt][l]{$\square$}\raisebox{.15ex}{\hspace{1em}}}               & \multicolumn{1}{l|}{\makebox[0pt][l]{$\square$}\raisebox{.15ex}{\hspace{1em}}}               & 54.9           & 77.3           & 4.22                & 59.9             \\ \hline
\multicolumn{1}{l|}{\makebox[0pt][l]{$\square$}\raisebox{.15ex}{\hspace{1em}}}               & \multicolumn{1}{l|}{\makebox[0pt][l]{$\square$}\raisebox{.15ex}{\hspace{1em}}}               & \multicolumn{1}{l|}{\makebox[0pt][l]{$\square$}\raisebox{.15ex}{\hspace{1em}}}               & \multicolumn{1}{l|}{\makebox[0pt][l]{$\square$}\raisebox{.15ex}{\hspace{1em}}}               & 53.2           & 76.1           & 4.20                & 54.6             \\ \bottomrule
\end{tabular}
}
\vspace{-2mm}
        
        \end{table}

\newpage
{\small
\bibliographystyle{ieee_fullname}
\bibliography{egpaper}
}

\end{document}